\documentclass[letterpaper]{article}
\usepackage{aaai}
\usepackage{times}
\usepackage{helvet}
\usepackage{courier}
\usepackage{graphicx}
\usepackage{booktabs}
\begin{document}
%
\title{Towards Quantification of Explainability in Explainable Artificial Intelligence Methods}
\author{Sheikh Rabiul Islam, William Eberle, and Sheikh K. Ghafoor\\
Tennessee Technological University\\
1 William L Jones Dr, Cookeville, TN 38505\\
}
\maketitle
\begin{abstract}
\begin{quote}
Artificial Intelligence (AI) has become an integral part of domains such as security, finance, healthcare, medicine, and criminal justice. Explaining the decisions of AI systems in human terms is a key challenge\textemdash{due to the high complexity of the model}, as well as the potential implications on human interests, rights, and lives . While Explainable AI is an emerging field of research, there is no consensus on the definition, quantification, and formalization of explainability. In fact, the quantification of explainability is an open challenge. In our previous work, we incorporated domain knowledge for better explainability, however, we were unable to quantify the extent of explainability. In this work, we (1) briefly analyze the definitions of explainability from the perspective of different disciplines (e.g., psychology, social science), properties of explanation, explanation methods, and human-friendly explanations; and (2) propose and formulate an approach to quantify the extent of explainability. Our experimental result suggests a reasonable and model-agnostic way to quantify explainability. 
\end{quote}
\end{abstract}

\noindent
\section{Introduction}
The use of Artificial Intelligence (AI) has spread into a spectrum of high-stakes applications from different domains such as security, finance, healthcare, medicine, and criminal justice, impacting human interests, rights, and lives. Following the pace of the demand, the successful AI-based models have also become so complex that their decisions have become too complicated to express in human terms. This further complicates their adoption in many sensitive disciplines, raising concerns from the ethical, privacy, fairness, and transparency perspectives. The root cause of the problem is the lack of explainability and/or interpretability of the decision. 

We have seen the very confusing use of the terms 'explainability' and 'interpretability' throughout the literature. Some treat these as the same and stick to one, while some differentiate among the two, and others use them ambiguously.  Research in Explainable Artificial Intelligence (XAI) is seeing a resurgence after three decades of slowed progress since some preliminary work on expert systems in the early 1980's. Currently, there is no consensus on the definition these terms. In addition, the quantification of explainability is another open challenge, which will be difficult until there is a consensus on the concrete definition of those terms (e.g., explainability, interpretability). 

In our previous work \cite{islam2019infusing}, we proposed an approach to incorporate domain knowledge in the model to make the prediction more explainable. However, we were unable to quantify the quality of explanations (i.e., validation of explainability). There has been some previous work  \cite{molnar2019quantifying} that mentions three notions for quantificiaiton of explainability. The first two notions involve experimental studies with humans that mainly investigate whether a human can predict the outcome of the model \cite{dhurandhar2017tip}, \cite{friedler2019assessing}, \cite{huysmans2011empirical}, \cite{poursabzi2018manipulating}, \cite{zhou2018measuring}. However, the third notion (proxy tasks) does not involve a human. Instead they use known truth as a metric (e.g., the less the depth of the decision tree, the more explainable the model). 

In this work, our contributions are as follows:
\begin{enumerate}
\item We define explainability and interpretability from a multi-domain perspective and clarify their resemblance (Section \ref{sec:explanation}).
\item We analyze the properties of explanation (Section \ref{subsec:properties_explanation}), explanation methods (Section \ref{subsec:properties_explanation_methods}), and human friendly explanations (Section \ref{subsec:properties_explanation_human}). 
\item We present a potential way to formalize and quantify explainability with experimental results (Section \ref{sec:methods} and \ref{sec:results}).
\end{enumerate}

\section{Explainability, and Interpretabiliy}\label{sec:explanation}
\cite{miller2018explanation} argue that most of the work on XAI focuses on the researcher's intuition of what constitutes a good explanation. However, there exists a vast area of research in philosophy, psychology, and cognitive science on how people generate, select, evaluate, and represent explanations and associated cognitive biases and social expectations towards the explanation process. Therefore, the author emphasizes that the research on Explainable AI should incorporate studies from these different domains.

From the social science perspective, according to \cite{miller2018explanation} and \cite{lombrozo2006structure}, explanation is both a process and product: 
\begin{itemize}
\item Cognitive explanation process: identifies the causes for an event, perhaps concerning particular counterfactual cases, and a subset of these causes is selected as the explanation.
\item Product: the explanation that results from the cognitive explanation process.  
\item Social process: the process of transferring knowledge between explainer and explainee, generally an interaction between groups of people, in which the goal is to provide the explainee with enough information to understand the causes of the event. 
\end{itemize}
This definition of explanation insists that it is both a process and product (i.e., outcome), and focuses on understanding the causes of the event. 

Furthermore, according to psychologist \cite{lombrozo2006structure}, ''Explanations are … the currency in which we exchanged belief''. This definition stresses the need for high fidelity for the explanation, in other words, gaining trust from the explanation recipient. Moreover, according to physicist \cite{deutsch1998fabric} who is the pioneer of quantum computation: \textit{explanations} consist of \textit{interpretations} of how the world works and why. This definition suggests that explanations are roughly equal to or a superset of interpretations. Interpretations are the building blocks of explanations; without interpretation, explanation is incomplete. Furthermore, the terms ''explainability'' and ''interpretability'' are simply the extent of explanation and interpretation accordingly. 

There are other mentionable definitions too. According to Miller (2017) \cite{miller2018explanation}, ''interpretability is the degree to which a human can understand the cause of a decision''. Besides, according to Kim \cite{kim2017interpretability} ''Interpretability is the degree to which a human can consistently predict the model's result''. Furthermore, according to \cite{ribeiro2016should} ''By 'explaining a prediction', we mean presenting textual or visual artifacts that provide qualitative understanding of the relationship between the instance's components (e.g. words in text, patches in an image) and the model's prediction''. In other words, \cite{ribeiro2016should} put an emphasis on the qualitative understanding of the relationship between input and output (i.e., a selective/suitable set of information pieces that together can refer to a cause to an event, in contrast to the complete causal attributions) . 

To this end, in the case of an intelligent system (i.e., AI/ML-based system), it is evident that explainability is more than interpretability in terms of importance, completeness, and fidelity of prediction. Based on that, we choose to keep our focus entirely on explainability instead of interpretability. Finally, analyzing different definitions from the literature, we come up with following:
\\[8pt]
\textbf{Definition}: \textit{Explainability} of an AI model's prediction is the extent of transferable \textit{qualitative understanding} of the relationship between model input and prediction (i.e., selective/suitable causes of the event) in a recipient friendly manner.
\\[8pt]
We proceed further based on this informal definition of explainability.

\section{Background}\label{sec:background}
There are two primary directions of research towards evaluation of explainability of an AI/ML model: (1) model complexity-based and (2) human study based.
\subsection{Model Complexity based Explainability Evaluation}
In the literature, model complexity and (lack of) model interpretability are often treated as the same \cite{molnar2019quantifying}. For instance, in \cite{furnkranz2012rule},  \cite{yang2017scalable}, model size is often used as a measure of interpretability (e.g., number of decision rules, depth of tree, number of non-zero coefficients). 

\cite{yang2017scalable} propose a scalable Bayesian Rule List (i.e., probabilistic rule list) consisting of a sequence of IF-THEN rules, identical to a decision list or one-sided decision tree. Unlike the decision tree that uses greedy splitting and pruning, their approach produces a highly sparse and accurate rule list with a balance between interpretability, accuracy, and computation speed. Similarly, the work of  \cite{furnkranz2012rule} is also rule-based. They attempt to evaluate the quality of the rules from a rule learning algorithm by: the observed coverage, which is the number of positive examples covered by the rule, which should be maximized to explain the training data well; and consistency, which is the number of negative examples covered by the rule, which should be minimized to generalize well to unseen data. 

Furthermore, according to \cite{ruping2006learning}, while the number of features and the size of the decision tree are directly related to the interpretability, the optimization of the tree size or features (i.e., feature selection) is costly as it requires generation of a large set of models and their elimination in subsequent steps. Besides, reducing the tree size (i.e., reducing complexity) increases error; however, they could not find a way to formulate the relation in simple functional form. 

More recently, \cite{molnar2019quantifying} attempt to quantify the complexity of the arbitrary machine learning model with a model agnostic measure. In that work, the author demonstrates that when the feature interaction (i.e., the correlation among features) increases, the quality of representations of explainability tools degrades. For instance, the explainability tool ALE Plot starts to show harsh lines (i.e., zigzag lines) as feature interaction increases. In other words, with more interaction comes a more combined influence in the prediction, induced from different correlated subsets of features (at least two), which ultimately makes it hard to understand the causal relationship between input and output, compared to an individual feature influence in the prediction. In fact, from our study of different explainability tools (e.g., LIME, SHAP, PDP), we have found that the correlation among features is a key stumbling block to represent feature contribution in a model agnostic way. Keeping the issue of feature interactions in mind, \cite{molnar2019quantifying} proposes a technique that uses three measures: number of features, interaction strength among features, and the main effect (excluding the interaction part) of features to measure the complexity of a post-hoc model for interpretation. Although, their work mainly focuses on model complexity for post-hoc models, that acted as a precursor to formulate our approach of explainability quantification. Our approach to quantify explainability is model agnostic and is for a model of any notion (e.g., pre-modeling, post-hoc).

\subsection{Human Study based Explainability Evaluation}
The following work deals with the application- and human-level evaluation of explainability involving human studies.

\cite{huysmans2011empirical} investigate the suitability of different alternative representation formats (e.g., decision tables, (binary) decision trees, propositional rules, and oblique rules) for classification tasks primarily focusing on the explainability of results rather than accuracy or precision. They discover that decision tables are the best in terms of accuracy, response time, confidence of answer, and ease of use. 

Furthermore, \cite{dhurandhar2017tip} argue that interpretability is not an absolute concept; rather, it is relative to the target model, and may or may not be relative to the human. Their finding suggests that a model is readily interpretable to a human when it uses no more than seven pieces of information \cite{miller1956magical}. Although, this might vary in task to task and person to person. For instance, a domain expert might consume a lot more detail information depending on their experience. 

The work of \cite{poursabzi2018manipulating} is a human-centered approach, focusing on previous work on human trust in a model from psychology, social science, machine learning, and human-computer interaction communities. In their experiment with human subjects, they vary factors (e.g., number of features, whether the model internals is clear or a black box) that makes a model more or less interpretable and measures how the variation impacts the prediction of human subjects. Their result suggests that participants who were shown a clear model with a small number of features were more successful in simulating the model's predictions and trusted the model's predictions. 

Furthermore, \cite{friedler2019assessing} investigate interpretability of a model based on two of its definitions: simulatability, which is a user's ability to predict the output of a model on a given input; and ''what if'' local explainability, which is a user's ability to predict changes in prediction in response to changes in input, given the user has the knowledge of a model's original prediction for the original input. They introduce a simple metric called runtime operation count that measures the interpretability, that is, the number of operations (e.g., the arithmetic operation for regression, the boolean operation for trees) needed in a user's mind to interpret something. Their findings suggest that the interpretability decreases with an increase in the number of operations. 


\section{Properties of Explanations and Explanation Methods}\label{sec:properties}

Although our main interest lies in explainable prediction, it depends on the model that generates the explanations. So, it is crucial to analyze the properties of both the explanation and the explanation method that generates the explanations. \cite{robnik2018perturbation} and \cite{molnar2018interpretable} attempt to define the properties of explanation methods and individual explanations, which we present in the following sections.

\subsection{Properties of Explanation Methods}\label{subsec:properties_explanation_methods}
Some properties of \textit{explanation methods} include:
\begin{itemize}
\item Expressive Power: Refers to the structure (e.g., decision trees, IF-THEN rules, weighted sum, and natural language) of the explanation method.
\item Translucency: The level in which the explanation method relies on looking into (e.g., coefficient for a linear regression, node splitting point in the tree-based approach) the ML model. For instance, explanation methods that rely on intrinsically interpretable methods like linear regression are highly translucent; however, counterfactual model agnostic explanation methods that leverage the changes in output in response to input for explanation have zero translucency as it does not look at model at all. The high translucency comes with the possible use of more information to generate the explanation. 
\item Portability: The number of ML models that the explanation method covers. Usually low translucency (i.e., ML model usually remains black box) comes with more portability. 
\item Algorithmic Complexity: The required time to generate the explanation.
\end{itemize}

\subsection{Properties of Individual Explanations}\label{subsec:properties_explanation}
Some properties of \textit{individual explanations} include:
\begin{itemize}
\item Accuracy: The explanation needs to be accurate enough when fidelity is essential. Although, low accuracy of explanation might be a cause of the low accuracy of the ML/AI model. 
\item Fidelity: How well the explanation approximates the prediction of the black-box model. Usually a highly accurate model with high fidelity leads to highly accurate explanations. Thus, fidelity and accuracy are related.
\item Consistency: Refers to the extent of consistency among explanations for different ML models on the same task. 	
\item Stability: Refers to the extent of similarity of explanations for similar instances. While consistency compares explanations between different models, stability compares the explanations of similar instances for a particular model.
\item Comprehensibility: The extent to which the recipient of explanation understands the explanation, which is very hard to measure. A few measures could be the number of features with non-zero weights in a linear model, or the number of rules in a tree. Usually a human can comprehend 7+-2 pieces of information at a time \cite{miller1956magical}
\item Certainty: Many ML models provide the probability of the target class. Similarly, explanations with a certainty value are expected to be useful. 
\item Degree of Importance: How well the explanation covers the important features or how well the explanation reflects parts of the explanations. For example, from the decision rules, we can understand which of the rules are more important (e.g., important features are at the beginning of a ruleset, or the top of the tree).
\end{itemize}

Recent laws \cite{goodman2016eu} \cite{algorithmic_accountability} focus on the impact of an algorithmic decision on human rights, interests, and lives. In that sense, we need a human-friendly explanation, which has slightly different properties than the explanation that does not consider humans as a key factor. 

\subsection{Properties of Human-friendly Explanation}\label{subsec:properties_explanation_human}
According to \cite{miller2018explanation}, humans usually prefer short explanations that contrast the current situation with a situation in which that event would not have occurred (a.k.a., counterfactual explanations). Sometimes abnormal causes provide good explanations. Besides, social context is also important for a good explanation as explanations are a social interaction among the explainer and the recipient of the explanation.  

Furthermore, the human friendly explanation does not consider all factors (i.e., selective in nature) for a particular prediction or behavior. However, if one needs to legally specify all influencing factors or need to debug the machine learning model, there is a need for a complete causal attribution \cite{molnar2018interpretable}, which is out of the scope of human-friendly explanation. That is the reason behind mentioning the term \textit{qualitative} instead of \textit{quantitative} in our definition of explanation (see Definition of Explainability).  

In summary, according to \cite{vstrumbelj2011general}, \cite{miller2018explanation} \cite {lipton1990contrastive}, and \cite{molnar2018interpretable}, human friendly explanations are:
\textit{contrastive}\textemdash{why a prediction (e.g., loan was rejected) was made instead of another prediction (e.g., loan was accepted)}, 
\textit{selected}\textemdash{does not cover the complete list of causes of an event}, 
\textit{social}\textemdash{social context (e.g., explaining to a layperson or domain expert) determines the nature of the explanations}, 
\textit{abnormal behavior focused}\textemdash{when a criteria (e.g., a particular feature value) is rare and has influence in the prediction, then it should be included (i.e., should have higher precedence in case of other criteria with the same influence) in the explanation} 
\textit{truthful}\textemdash{explanation should be as truthful as possible, although selectiveness comes first which might exclude some of the true reasons}, 
\textit{consistent}\textemdash{consistent with prior belief}, and 
\textit{general and probable}\textemdash{good explanations are general and probable, although this contradicts with the claims that abnormal causes make good explanations, abnormal causes have higher preference over general and probable explanations (see \textit{abnormal behavior focused} point)}.

\section{Explainability Quantification Method}\label{sec:methods}
Usually, humans can relate and process 7+-2 pieces of information (i.e., cognitive chunks) to understand something \cite{miller1956magical}. For instance, suppose that, in the most generalized form, the quality of an explanation is dependent upon the number of cognitive chunks that the recipient has to relate to in order to understand an explanation (i.e., the less, the better). 
Lets assume, E {=} explainability; N\textsubscript{c} {=} number of cognitive chunks; I {=} interaction; N\textsubscript{i} {=} number of input cognitive chunks; and N\textsubscript{o} {=} number cognitive chunks involved in the explanation representation (i.e., output cognitive chunks).
\begin{equation} \label{eq:formula1}
E = \frac{1}{N_{c}} 
\end{equation}
However, sometimes, these cognitive chunks are correlated and their influence/contribution/abilities are not mutually exclusive. This interaction among cognitive chunks complicates the explainability. So we penalize Formula \ref{eq:formula1} for having an interaction among cognitive chunks, resulting in Formula \ref{eq:formula2}. 
\begin{equation} \label{eq:formula2}
E = \frac{1}{N_{c}} + (1 - I)
\end{equation}
Where, the interaction \textit{I} ranges in between 0 and 1, and the less the interaction, the better the explainability, so we take the complement of that.

Formula \ref{eq:formula2} is in a form that can be applied to any of the application, domain, or proxy level explanation evaluations described before. However, from the perspective of an application and domain level evaluation of explainability, to progress further, we need human studies that are out of the scope of this work. Instead, in this work, we focus on the proxy level evaluation of explainability that considers different properties of output representation (e.g., depth of decision tree, length of rule list) as a metric for evaluation. 

Furthermore, we need to breakdown the number of cognitive chunks more to get a better evaluation. Both the number of input cognitive chunks in the model and the number of output cognitive chunks involved in the representation of output are important to understand the causal relationship, which is vital for explanation.  While the ideal explainability case would be when there is only one input and one output cognitive chunk (no chance of interaction), that is unusual in real-world situations. Following the segregation of input and output cognitive chunks, Formula \ref{eq:formula2} can be re-written as Formulas \ref{eq:formula3} and \ref{eq:formula4}:

\begin{equation} \label{eq:formula3}
E = \frac{1}{N_{c}} + (1 - I)
\end{equation}
\begin{equation} \label{eq:formula4}
E = \frac{1}{N_{i}} + \frac{1}{N_{o}} + (1 - I)
\end{equation}

where N\textsubscript{i} refers to the number of input cognitive chunks
and N\textsubscript{o} refers to the number cognitive chunks involved in the explanation representation (i.e., output cognitive chunks). Usually, the more these cognitive chunks, the more complicated the explanation becomes. So, the ratio of the best case (i.e., one cognitive chunk) and observed case is added towards total explainability. 

Also, Formula \ref{eq:formula4} has three predicates, which might have different influences on the quantification of explainability, and different domains might have different implications (e.g., accuracy vs explainability trade off). So, we add a weight term with each of the predicates, considering the weights are constant (e.g., .3333) by default, and their summation is equal to 1: w\textsubscript{1} + w\textsubscript{2} + w\textsubscript{3} = 1. However, these weights can be set to a different distribution, perhaps dependant upon a particular domain (e.g., healthcare, finance).

After the addition of the weight terms, Formula \ref{eq:formula4} becomes Formula \ref{eq:formula5}:
\begin{equation} \label{eq:formula5}
E = \frac{w_{1}}{N_{i}} + \frac{w_{2}}{N_{o}} + w_{3} (1 - I)
\end{equation}
Formula \ref{eq:formula5} can then be used to quantify the explainability  of the explanation method (i.e., global explainability). We can use Formula \ref{eq:formula5} to also quantify the explainability of an instance level prediction (i.e., local explainability). In that case, the first predicate of Formula \ref{eq:formula5}  (including the weight term) remains the same (i.e., the same number of input chunks). However, predicate 2 and predicate 3 will be different from instance to instance as a different set of cognitive chunks with different interaction strengths might be involved in the representation of explanation for a particular instance as explanations are selective.

\section{Experimental Results}\label{sec:results}
We provide a brief overview of the experimental settings of our previous work \cite{islam2019infusing} that we will then use for our proposed explainability quantification method. In this example, we incorporate domain knowledge in the model for mortgage bankruptcy prediction, and apply different supervised ML algorithms in three different ways:
\begin{enumerate}
\item Using \textit{original features}:  we use all original features that have a non-zero effect on prediction (features with zero effect were removed in a data pre-processing stage by observing the Random Forest's feature importance). 
\item Using \textit{domain-related features}:  we select a subset of features that match with extracted domain knowledge (e.g., 5 C's of credit) from the domain. Traditionally, credit risk is assessed using the 5C's of credit (character, capital, capacity, collateral, and cycle), a popular domain principle to for determining credit risk. We extract necessary domain knowledge and map to the 5 C's of credit to get the domain features\cite{islam2019infusing}. 
\item Using \textit{newly constructed features}:  we construct a very concise set of new features (i.e., five features\textemdash{one feature for each element of the domain principle 5C's of credit}) from the domain-related features using the quantitative measure of gain/compromises (i.e., the cumulative sum of related feature values times the correlation coefficient) associated with each element of the domain principle. 
\end{enumerate}

\begin{figure}[h]
  \centering
  \includegraphics[width=.99\linewidth]{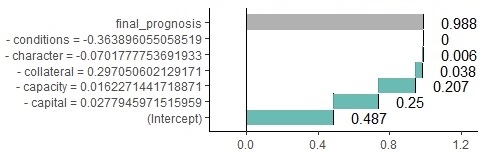}
  \caption{Breakdown of feature contributions for a random sample.}
    \label{fig:breakdown}
\end{figure}

Furthermore, we represent the predicted output as a composition of individual elements of the domain principle 5C's of credit(i.e., \textit{newly constructed features}) (Figure \ref{fig:breakdown}).

One will notice that the representation (Figure \ref{fig:breakdown}) of the prediction in terms of the \textit{newly constructed features} provides better explainability as the final prediction is segregated into the individual influences of a very concise set of intuitive features (i.e., five compared to 30). However, there is no way to quantify the level of perceived explainability. To use our proposed formula in this work (Formula \ref{eq:formula5}) to quantify explainability, we need to calculate the \textit{interaction strength} (\textit{I}) too. We measure the interaction strength among features using R's iml package that uses the partial dependence of an individual features as the basis for calculating  \textit{interaction strength} (\textit{I}). 

\begin{table}
\centering
\caption{Comparison of explainability}
\label{tab:explainability}
\begin{tabular}{llll}
\toprule
                           & Original & Domain & Constructed  \\
\midrule
Input chunks (N\textsubscript{i})               & 30                & 7                       & 7                           \\
Output chunks (N\textsubscript{o})              & 30                & 7                       & 5                           \\
Int. Strength (I)          & 0.556             & 0.5233                  & 0.5251                      \\
Explainability (E)            & 0.1701            & 0.2539                  & 0.2723                      \\
\bottomrule
\end{tabular}
\end{table}


\begin{figure}[h]
  \centering
  \includegraphics[width=.99\linewidth]{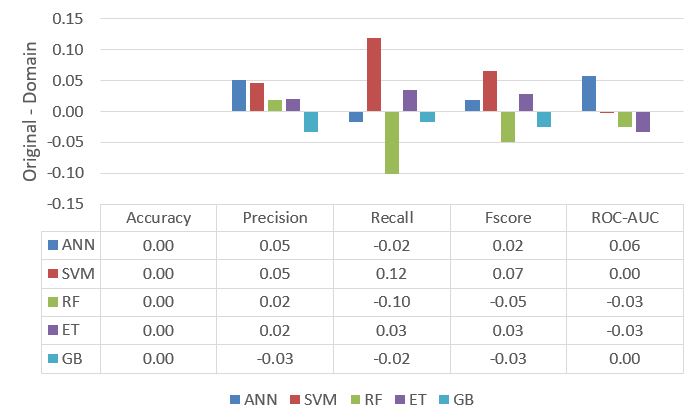}
  \caption{Dispersion in performance\textemdash{\textit{original features} minus \textit{domain-related features}}}
    \label{fig:original_vs_domain}
\end{figure}

\begin{figure}[h]
  \centering
  \includegraphics[width=.99\linewidth]{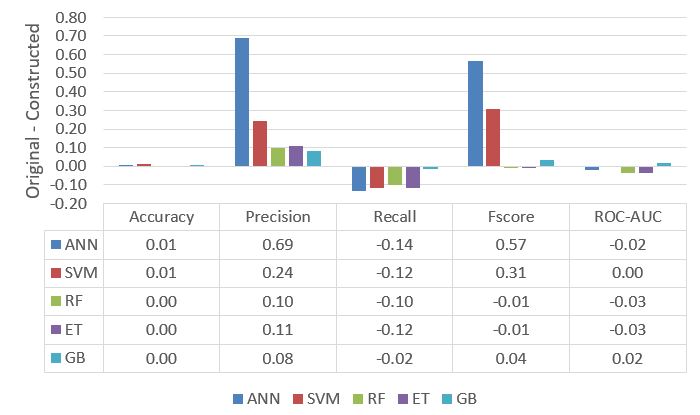}
  \caption{Dispersion in performance\textemdash{\textit{original features} minus \textit{newly constructed features}}}
    \label{fig:original_vs_constructed}
\end{figure}

Applying Formula \ref{eq:formula5}, on metadata (Table \ref{tab:explainability}) of three different feature settings, we see that \textit{newly constructed features} (5'C of credit) provide the best explainability score of .2723, which is an improvement of 60.14\% compared to the  0.17 that we get using the \textit{original features} (Table \ref{tab:explainability}). In fact, even if we apply the state-of-the-art methods of post-hoc interpretability/explainability like SHAP, the explainability will be still limited to 0.17 as it does not reduce the number of cognitive chunks to represent output. Besides, using \textit{domain related features}, the explainability score is .2538, which is better than using the original features, although a little worse than using the \textit{newly constructed features}. 

Furthermore, this explainability gain comes with a negligible cost of performance (Figure \ref{fig:original_vs_domain}, Figure \ref{fig:original_vs_constructed}) which is expected and a known trade off for explainability methods. In fact, for a few algorithms (e.g., Random Forest (RF), Gradient Boosting (GB)), \textit{the newly constructed features}, and \textit{domain related features} lead to better \textit{recall}, which is crucial for anomaly detection where the target class instances are very few compared to non-target class instances. In that sense, besides a better explainability, the domain knowledge benefits us in terms of different performance (e.g., better recall, less features leads to less computation time) measures too.   

\section{Conclusion}
Explainable decisions from commercial AI systems are going to be a standard imposed by regulators to eliminate bias and discrimination, and ensure trust. Our work attempts to establish a concrete definition of explainability, properties of explanations and explanation methods, and a way to quantify explainability which is transferable to a variety of explainability methods or tools. For the quantification of explainability, we only present an approach for the proxy method. In the future, we would like to address human studies as an extension of this work.

\bibliography{bib.bib}
\bibliographystyle{aaai}
\end{document}